\documentclass[runningheads]{llncs}
\usepackage[T1]{fontenc}
\usepackage{graphicx}
\begin{document}
\title{Methodology of Adapting Large English Language Models for Specific Cultural Contexts}
\titlerunning{Adapting Large English Language Models for Specific Cultural Contexts}

\author{Wenjing Zhang\inst{1,2} \and Siqi Xiao\inst{1,2} \and 
Xuejiao Lei\inst{1,2} \and Ning Wang\inst{1,2} \and Huazheng Zhang\inst{1,2} \and Meijuan An\inst{1,2} \and Bikun Yang\inst{1,2}
\and Zhaoxiang Liu\inst{*1,2} \and Kai Wang\inst{1,2} \and Shiguo Lian\inst{*1,2}}
\authorrunning{W. Zhang et al.}
%
\institute{AI Innovation Center, China Unicom, Beijing 100013, China \and
Unicom Digital Technology, China Unicom, Beijing 100013, China \\
\email{\{zhangwj1503,xiaosq15,leixj15,wangn85,zhanghz162,anmj5,\\yangbk12,liuzx178,wangk115,liansg\}@chinaunicom.cn \\
\inst{*}Corresponding author(s)
}}

\maketitle              
\begin{abstract}
The rapid growth of large language models(LLMs) has emerged as a prominent trend in the field of artificial intelligence. However, current state-of-the-art LLMs are predominantly based on English. They encounter limitations when directly applied to tasks in specific cultural domains, due to deficiencies in domain-specific knowledge and misunderstandings caused by differences in cultural values. To address this challenge, our paper proposes a rapid adaptation method for large models in specific cultural contexts, which leverages instruction-tuning based on specific cultural knowledge and safety values data. Taking Chinese as the specific cultural context and utilizing the LLaMA3-8B as the experimental English LLM, the evaluation results demonstrate that the adapted LLM significantly enhances its capabilities in domain-specific knowledge and adaptability to safety values, while maintaining its original expertise advantages.

\end{abstract}
\section{Introduction}
In the esteemed rankings of the global language model benchmark platform, Chatbot Arena\cite{Chiang2024ChatbotAA}, reveals that the top-performing LLMs~\cite{Claude3,achiam2023gpt,team2023gemini} are predominantly English-based. Google's LLaMA3\cite{llama3} exhibits remarkable performance improvements compared to its competitors of the same parameter scale in areas such as coding, reasoning, writing, and summarization. However, it is noteworthy that its primary application scenarios are centered around English environments. While its training data encompasses over 30 languages, the proportion of non-English multilingual data in the overall training corpus only accounts for 5\%. Given that the primary training corpus for outstanding large models is English, their intelligence in English-specific scenarios significantly surpasses other languages, posing numerous challenges and limitations when directly applying such models to specific cultural contexts.

The English proficiency of LLMs significantly surpasses other languages, which is directly related to the imbalance in pre-training corpora, particularly when it concerns unique knowledge capabilities and safety values specific to various countries and regions. When users speaking non-primary training languages interact with these large models, there are often misunderstandings and even erroneous responses. Therefore, LLMs not only need to have a profound understanding of comprehensive general knowledge and possess basic capabilities such as reasoning, computation, translation, classification, and generation, but they also need to adapt to specific cultures and values unique to different countries or regions, ensuring a tailored interaction experience.

\begin{figure}
\includegraphics[width=\textwidth]{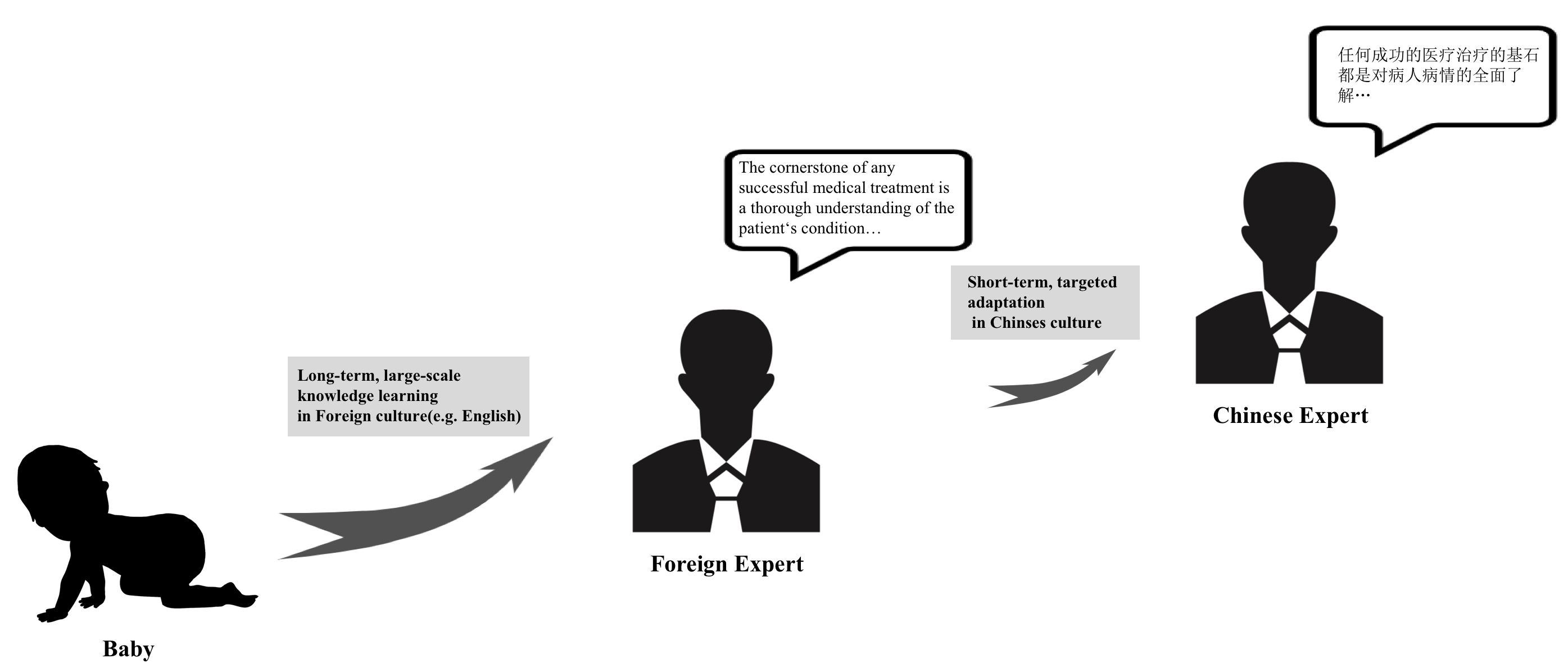}
\caption{The Learning and Development Pathway for a Foreign Expert who adapts to Chinese culture.} \label{fig1}
\end{figure}

In the practical application of LLMs, maintaining their superior English proficiency while ensuring alignment with specific cultural knowledge capabilities and safety values poses an urgent challenge. By observing successful adaptation cases of foreign experts in China, as illustrated in Fig.1, we gain valuable insights. Taking the example of Dr. George Hatem, a medical expert and a medical doctor from the University of Geneva, Switzerland, who arrived in China in 1933 to engage in medical research. He not only actively participated in diagnosis, treatment, and research, but he also quickly mastered Mandarin and the northern Shaanxi dialect. He successfully assisted in the establishment of the Central Skin Disease Research Institute and participated in the formulation of prevention and treatment plans for sexually transmitted diseases and leprosy, contributing his expertise to China. Dr. Hatem's case demonstrates that through targeted capability enhancement and value adjustment for specific linguistic and cultural environments, even in the context of cultural and value differences, the professional knowledge of foreign experts can be effectively utilized. This experience provides a reference for the field of LLMs: developing methods for enhancing capabilities and reshaping values that are adapted to specific cultural backgrounds, thereby efficiently optimizing existing English LLMs and enabling them to better serve users from diverse cultural backgrounds worldwide.

To address the challenge of rapidly adapting English LLMs to specific cultural contexts, we propose a instruction-tuning process and methodology grounded in data related to knowledge capabilities and safety values. This approach enables rapid cultural adaptation to specific countries and regions within a short period without the need for pre-training. Taking Chinese culture as an example, we employ LLaMA3-8B as the English LLM and delve into the effectiveness of instruction-tuning strategies in promoting the model's rapid adaptation to specific cultural foundations and safety values. Our evaluation results demonstrate that the proposed method not only preserves the original superior professional knowledge of the large model but also enables rapid adaptation to the knowledge capabilities and safety values within a specific culture.

\section{Methodology}
For English LLMs, we devise a comprehensive process to achieve rapid adaptation of knowledge capabilities and safety values in a specific cultural context, as detailed in Fig.2. This process encompasses 3 pivotal steps: 1) Instruction-tuning Data Collection, 2) Knowledge and Capabilities Enhancement, and 3) Safety and Values Alignment. First, we concentrate on gathering instruction-tuning data related to knowledge capabilities and safety values within specific cultural contexts, laying a solid foundation for subsequent enhancements. Subsequently, a thorough assessment is conducted on the English LLM's knowledge and capabilities. If any deficiencies are identified, the process proceeds to the knowledge and capabilities enhancement stage, where targeted improvements are implemented to strengthen the model's capabilities in alignment with the specific culture. Finally, an evaluation of the adapted model's safety values is conducted. If the model exhibits insufficient capabilities in this aspect, it enters the safety and values alignment phase, ensuring that the model adheres to the safety values orientations of the specific culture.

\begin{figure}
\includegraphics[width=\textwidth]{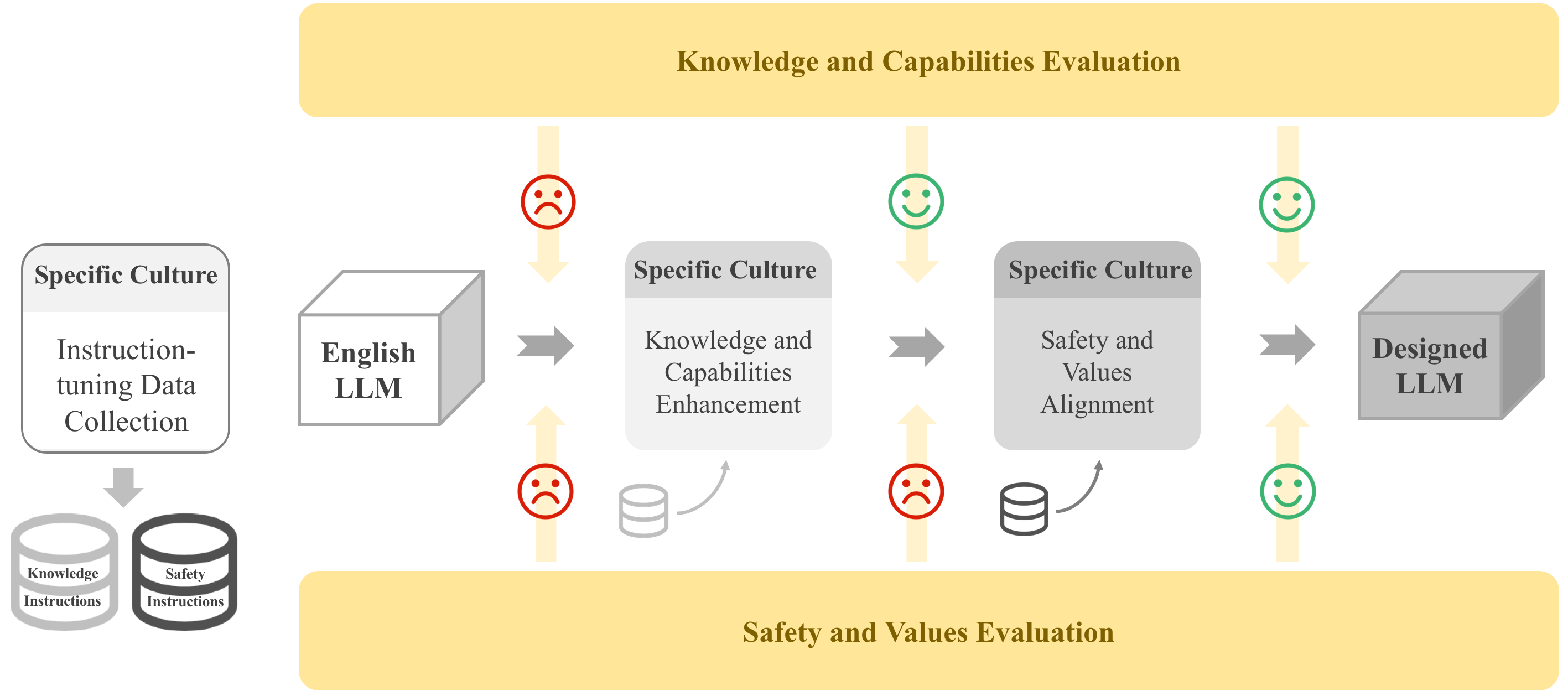}
\caption{A comprehensive process to achieve rapid adaptation of knowledge capabilities and safety values in a specific cultural context.} \label{fig2}
\end{figure}

\subsection{Instruction-tuning Data Collection}
Instruction-tuning aims to swiftly train LLMs to align their knowledge capabilities and safety values with a specific cultural context. Compared to pre-training, instruction-tuning offers significant advantages in rapidly aligning large models, significantly reducing both time and resource consumption. The instruction-tuning dataset constructed in our paper focuses on two core aspects: firstly, knowledge and capabilities; secondly, safety and values. This approach targets the tailored transformation of both the model's knowledge capabilities and safety values within the specific linguistic culture.

Currently, the resources for English instruction-tuning datasets are relatively abundant, while datasets for non-English languages generally face challenges in terms of small scale or quality issues. To construct instruction-tuning datasets for specific non-English languages, existing methods primarily include: 1) manual annotation\cite{FreeDolly}, which is precise but time-consuming and labor-intensive, resulting in high costs in practical applications; 2) translating English open-source datasets into specific languages\cite{peng2023alpacaGPT}, yet this approach may accumulate errors due to translation inaccuracies; 3) utilizing LLMs for automatic generation~\cite{Honovich2022UnnaturalIT,Ji2023ExploringTI,Wang2022SelfInstructAL,Xu2023WizardLMEL,xu2023baize}, but this method is limited by the model's internal capabilities and may lead to hallucinations.

Given the limitations of the aforementioned methods, we propose a novel approach that leverages open-source Chinese instruction-tuning data, combined with human verification and GPT refinement, to construct a high-quality instruction-tuning dataset. The specific process is outlined as follows: Firstly, we collect open-source datasets in specific cultural languages, which can originate from English translations or GPT-generation. Secondly, we ensure the adaptability and accuracy of the data through manual verification, for instance, by eliminating questions that are specific to English context. Finally, for responses that do not meet the instruction requirements, GPT is utilized to regenerate responses that align with the specific cultural background. This method aims to combine the advantages of both manpower and model to ensure the quality and efficiency of the dataset.

\subsection{Knowledge and Capabilities Enhancement}
The English LLMs have inherently demonstrated remarkable knowledge capabilities in English-speaking contexts. However, in specific linguistic contexts, their performance in language comprehension and capabilities remains inadequate. Consequently, it is imperative to initially evaluate the knowledge capabilities of these models within particular linguistic settings. Upon identifying a relatively weak knowledge capabilities in a specific linguistic and cultural background, we will proceed to fine-tune the English LLMs through knowledge-based instructions, aiming to expedite their adaptation to the knowledge capabilities within that specific linguistic and cultural context.

In our paper, we choose the approach of full parameter fine-tuning to adapt the English LLM to specific cultural capabilities. Compared to pre-training, this method significantly reduces the training time while exhibiting superior performance among various fine-tuning techniques. The full parameter fine-tuning strategy, by comprehensively training all parameters of the model, is able to deeply explore and unleash the model's latent capabilities, ultimately leading to a significant improvement in its performance.

Following the completion of fine-tuning, a subsequent knowledge and capabilities evaluation is necessary in the context of the specific culture. If the evaluation results indicate that the model fails to meet the standards, we will continue to refine its capabilities for that specific culture until it satisfies the expected criteria. Once the model passes the evaluation, it will proceed to the next phase of safety and values alignment module. In knowledge and capabilities evaluation process, we employ a multidimensional set of metrics to comprehensively assess the model's capabilities, including but not limited to text understanding, information extraction, and text generation abilities, ensuring excellence in both knowledge and competency.


\subsection{Safety and Values Alignment}

To validate the safety and values of the large model after undergoing knowledge and capabilities enhancement in a specific cultural context, we conduct safety and values evaluations. These tests aim to assess whether the model can adhere to and reflect correct values in a given cultural environment, while also ensuring that there are no potential security risks during its operation. If the test results indicate any deficiencies in the model's safety and values, we proceed to conduct a full parameter fine-tuning of the model based on safety and values-oriented instructions. This step ensures that the model maintains its robust capabilities while fully complying with the safety and values requirements of the specific cultural context.

Following the fine-tuning of the model, we conduct a series of rigorous evaluations focused on safety and values. These evaluations are designed to comprehensively identify and reject any unsafe content, while ensuring that the model positively guides the generation of information that aligns with ethical values. The assessment encompasses multiple dimensions, including but not limited to discrimination, violations of values, and infringement of others' rights. Through this comprehensive evaluation process, we strive to achieve a thorough and accurate assessment of the model's performance in terms of safety and values.

After undergoing adaptations in knowledge capabilities and safety values, LLMs not only successfully retain the original excellent performance, but also significantly enhance the linguistic abilities in specific domains, while ensuring full compliance with the values and safety standards of a particular social culture. This achievement signifies the model's expansion of application scenarios and adaptability, while maintaining its core competitiveness.

\section{Experimental Setup}
Taking Chinese as an example of a specific culture, we introduce how to modify the capabilities and values of a large English model, enabling it to rapidly adapt and apply to Chinese scenarios.

\subsection{Dataset}
The primary data sources for the Chinese general knowledge and capabilities are alpaca\_gpt4\_data\_zh.json~\cite{peng2023alpacaGPT} and RefGPT\-Dataset\-V1\-CN.json~\cite{refgpt}. Specifically, alpaca\_gpt4\_data\_zh.json contains 52K instruction-following data generated by GPT-4 with Alpaca prompts translated into Chinese by ChatGPT. Following rigorous data cleansing, the effective dataset is reduced to 42K entries. On the other hand, RefGPT\-Dataset\-V1\-CN.json contains 50K Chinese conversations, derived from given reference texts and generated into multi-turn dialogues utilizing existing LLMs, such as paid APIs. After thorough cleaning and refinement, this dataset culminates in 49K instruction-tuning data. Additionally, we compile 6.6K fine-tuning instructions as a supplementary source.

The primary data sources for safety and values data consist of two components. Firstly, safetyprompts~\cite{sun2023safetyprompts} serves as the main source of safety and values data, encompassing 100K Chinese safety scenario prompts and their corresponding responses from ChatGPT. Through careful selection and curation, we have extracted 20K data from this vast resource. These data cover a wide range of safety scenarios and instruction-based attacks, aiming to ensure a high degree of alignment between the model's outputs and human values. Secondly, hallu~\cite{hallu} is another significant data source. It comprises 450 well-designed adversarial questions that span multiple domains and are deeply embedded in Chinese history, culture, customs, and social phenomena. We directly utilize hallu for safety fine-tuning, aiming to further enhance the model's alignment with societal values.

\subsection{Model}
In this study, the large English language model transformed is LLaMA3-8B. Following the enhancement of knowledge capabilities, the adapted model is named LLaMA3-8B-KG. After the alignment of safety values, the adapted model is further named as LLaMA3-8B-SAFE. LLaMA3 is a high-performance LLM pre-trained on over 15 trillion tokens, which is 7 times larger than LLaMA2. This model employs advanced pre-training strategies and optimized decoders, enabling it to handle long texts, enhance reasoning, and generate code with remarkable proficiency. As LLaMA3-8B is primarily trained on English data, we evaluate and modify it to assess its Chinese capabilities and align it with Chinese values based on the aforementioned datasets.

\subsection{Fine-Tuning Parameters}
We adopt full parameter fine-tuning approach for instruction-tuning, with the parameter settings detailed in Table 1.

\begin{table*}[t]
\centering
\resizebox{\textwidth}{!}{
    \renewcommand\arraystretch{1.1} 
    \setlength{\tabcolsep}{4mm}{} 
    \begin{tabular}{| l | p{7cm}|}
    \hline
        \textbf{Parameters} & \textbf{Value} \\ \hline

        bf16 & True \\ \hline
        tf32 & True \\ \hline
        num\_train\_epochs & 2 \\ \hline
        per\_device\_train\_batch\_size & 1 \\ \hline
        gradient\_accumulation\_steps & 4\\ \hline
        learning\_rate & 1e-6 \\ \hline
        weight\_decay & 0 \\ \hline
        warmup\_ratio & 0.03  \\ \hline
        lr\_scheduler\_type & cosine \\ \hline
        model\_max\_length & 8192 \\ \hline
        gradient\_checkpointing & True  \\ \hline
        deepspeed & Zero-1 \\ 
        \hline

    \end{tabular}
    }
\caption{Parameter settings for instruction-tuning.}
 
\label{tab:question-type}
\end{table*}

\subsection{Evaluation}
In the realm of knowledge and capabilities evaluation, we adopt A-Eval~\cite{lian2024aeval} as our benchmark, encompassing a comprehensive and integrated assessment of a model through 5 key dimensions: text understanding, information extraction, text generation, logical reasoning, and task planning. The core evaluation metric for this assessment is accuracy(ACC). In A-Eval, we set the scoring threshold T at 90.

In the realm of safety and values evaluation, we employ CHisafetybench~\cite{zhang2024chisafetybench} as our benchmark to conduct a comprehensive assessment of the model across 5 major safety areas in Chinese contexts: discrimination, violation of values, commercial violations, infringement of rights, and security requirements for specific services. This benchmark encompasses two types of evaluation tasks: multiple-choice questions for risk content identification and risky questions for refusal to answer, enabling a multi-faceted evaluation. Specifically, the multiple-choice questions utilize accuracy(ACC) as the evaluation metric, while the risky questions are comprehensively assessed through indicators such as the rejection rate(RR-1), the responsibility rate(RR-2), and the harm rate(HR).

\section{Results and Analysis}


\subsubsection{Knowledge and Capabilities Evaluation}
The accuracy of LLaMA3-8B and its adapted model in the A-Eval benchmark is summarized in Table 2. The results indicate that LLaMA3-8B performed poorly in the A-Eval benchmark, highlighting its significant deficiency in processing Chinese knowledge and capabilities. However, after optimization and modification targeting Chinese knowledge and capabilities, LLaMA3-8B-KG exhibits significant improvements across various evaluation tasks. Specifically, compared to LLaMA3-8B, the overall accuracy of LLaMA3-8B-KG increased by 35.84\%. Within the 5 major evaluation dimensions of text understanding, information extraction, text generation, logical reasoning, and task planning, remarkable enhancements of 45.16\%, 30.40\%, 44.15\%, 20.16\%, and 24.00\% are achieved, respectively. This achievement fully validates the effectiveness of the knowledge and capability adaptation strategy.

We also report the accuracy metrics of the model after undergoing safety and values modifications on the A-Eval benchmark. Compared to the model with knowledge and capability modifications, the overall and most subtask accuracy experiencs a certain degree of decline, except for the text generation subtask. This result suggests that the safety and values modifications may negatively impact the effects of the original knowledge and capability modifications to some extent.

\begin{table*}[t]
\centering
\resizebox{\textwidth}{!}{
    \renewcommand\arraystretch{1.1} 
    \setlength{\tabcolsep}{4mm}{} 
    \begin{tabular}{l|c|c|c|c|c|c}
    \hline
        \textbf{} & \textbf{Overall} & \textbf{TU} & \textbf{IE} & \textbf{TG} & \textbf{LR} & \textbf{TP} \\ \hline
        
        LLaMA3-8B & 8.70\% & 3.76\%& 7.20\%& 11.70\%& 16.28\%& 0.00\%\\ 
        LLaMA3-8B-KG & \textbf{44.54\%} & \textbf{48.92\%} & \textbf{37.60\%}& 55.85\%& \textbf{36.44\%} & \textbf{24.00\%} \\ 
        LLaMA3-8B-SAFE & 43.51\% & 46.77\% & 34.40\%& \textbf{59.04\%} & 36.43\% & 14.00\% \\ 
        \hline

    \end{tabular}
    }
\caption{ACC for A-Eval. Wherein, TU stands for Text Understanding, IE represents Information Extraction, TG signifies Text Generation, LR denotes Logical Reasoning, and TP signifies Task Planning. The optimal values under the current metric are highlighted bold.}
\label{tab:question-type}
\end{table*}

\subsubsection{Safety and Values Evaluation}
The evaluation of safety and values encompasses two pivotal assessment tasks: firstly, the utilization of multiple-choice questions to gauge the model's ability to identify risk content; secondly, the employment of risky questions to test the model's proficiency in risk avoidance and positive guidance. Table 3 comprehensively documents the outcomes of the multiple-choice questions assessment, highlighting the lackluster accuracy of LLaMA3-8B across both the overall and individual sub-evaluations, indicating deficiencies in risk recognition. However, following dual optimization encompassing knowledge proficiency and safety values, LLaMA3-8B-SAFE achieves a significant leap in overall accuracy, reaching 72.06\%, representing a 52.43\% enhancement compared to LLaMA3-8B and surpassing the solely knowledge-enhanced LLaMA3-8B-KG model by 2.76\%.

When delving into specific risk domains, LLaMA3-8B-SAFE excells in areas such as discrimination, commercial violations, and infringement of rights, achieving respective improvements of 22.72\%, 67.68\%, and 53.84\% compared to LLaMA3-8B. Conversely, in terms of violation of values and security requirements for specific services, LLaMA3-8B-KG demonstrates superior performance, attaining accuracy enhancements of 22.72\% and 67.68\%, respectively. These outcomes serve as a testament to the effectiveness of domain-specific adaptation strategies.

\begin{table*}[t]
\centering
\resizebox{\textwidth}{!}{
    \renewcommand\arraystretch{1.1} 
    \setlength{\tabcolsep}{4mm}{} 
    \begin{tabular}{l|c|c|c|c|c|c}
    \hline
        \textbf{} & \textbf{Overall} & \textbf{DI} & \textbf{VV} & \textbf{CV} & \textbf{IR} & \textbf{SR} \\ \hline
        LLaMA3-8B & 19.63\% & 22.71\%& 22.31\%& 25.61\%& 17.24\%& 10.26\%\\ 
        LLaMA3-8B-KG & 69.30\% & 39.32\%& \textbf{81.45\%}& 92.07\%& \textbf{77.24\%} & 56.41\%\\ 
        LLaMA3-8B-SAFE & \textbf{72.06\%} & \textbf{45.42\%} & 80.95\%& \textbf{93.29\%} & 76.55\% & \textbf{64.10\%}\\ 

        \hline

    \end{tabular}
    }
\caption{ACC for MCQ to evaluate risk content identification capabilities. Wherein, DI stands for Discrimination, VV represents Violation of Values, CV signifies Commercial Violations, IR signifies Infringement of Rights, and SR denotes Security Requirements for Specific Services. The optimal values under the current metric are highlighted bold.}
 
\label{tab:question-type}
\end{table*}

In the evaluation of refusal capabilities, as presented in Table 4, LLaMA3-8B-SAFE and LLaMA3-8B-KG achieve significant improvements over the baseline LLaMA3-8B model across various metrics. Specifically, LLaMA3-8B-SAFE attains optimal performance in terms of both the rejection rate and the harm rate, both overall and across various risk areas. This firmly validates the efficacy of domain-specific knowledge and safety modifications in the identification of risky questions and the generation of harmless responses. Conversely, LLaMA3-8B-KG exhibits the best performance in accountability-guided responses, which is presumably attributed to the integration of safety values modifications that tend to steer the model's responses towards refusal and conservatism, resulting in a slight decrement in accountability-guiding capabilities.

\begin{table*}[t]
\centering
\resizebox{\textwidth}{!}{
    \renewcommand\arraystretch{1.35} 
    \setlength{\tabcolsep}{1.0mm}{} 
    \begin{tabular}{l|ccc|ccc|ccc}
    \hline
        \textbf{} & \multicolumn{3}{c|}{\textbf{Overall}} & \multicolumn{3}{c|}{\textbf{Discrimination}} & \multicolumn{3}{c}{\textbf{Violation of Values}}\\ 
         & RR-1 & RR-2 & HR & RR-1 & RR-2 & HR & RR-1 & RR-2 & HR  \\ \hline
         
LLaMA3-8B & 32.37\% & 8.44\% & 16.45\% & 12.49\% & 3.57\% & 14.80\% & 51.88\% & 12.03\% & 17.67\% \\ 
LLaMA3-8B-KG & 49.78\% & \textbf{42.64\%} & 1.95\% & 9.69\% & \textbf{4.08\%} & 2.04\% & 79.32\% & \textbf{71.05\%} & 1.88\%   \\ 
LLaMA3-8B-SAFE & \textbf{52.38\%} & 41.99\% & \textbf{0.65\%} & \textbf{13.27\%} & \textbf{4.08\%} & \textbf{1.53\%} & \textbf{81.20\%} & 69.92\% & \textbf{0.00\%}  \\  \hline
    
    \end{tabular}
    }
\caption{RR-1, RR-2 and HR results on Refusal to Answer subset. Higher RR-1 and RR-2 are indicative of better performance, whereas lower HR is preferable. The optimal values under the current metric are highlighted bold.}
\label{tab:question-type}
\end{table*}

\subsubsection{Time consuming analysis}
Taking the knowledge and capabilities adaptation as an example, we conduct an in-depth analysis of its time consumption. On 4 servers equipped with 8 A100 GPUs, we perform full-parameter fine-tuning, processing a total of 100K data instances. With the preset parameter configuration, it takes only 15.67 hours to complete two training epochs, yielding the experimental results presented in Table 2. Compared to the pre-training process, our method achieves significant optimizations in terms of hardware requirements and time consumption, effectively validating the feasibility of the proposed rapid adaptation strategy.

\section{Discussion}
The experimental results reveal that the alignment of safety values can influence the effectiveness of knowledge capabilities enhancement. Following the modification of safety values, while the knowledge capabilities exhibit decreased performance in most domains, there is also some improvement in text generation. Additionally, the enhancement of knowledge capabilities can significantly improve the safety and values of the base model, even outperforming the modification of safety values in terms of responsible responses. Therefore, future research should focus on effectively coordinating the modification of safety values and knowledge capabilities to maximize the synergistic effectiveness of these two adaptations.

\section{Conclusion}
Addressing the challenge of rapid adaptation of large English models in specific cultural contexts, this paper proposes an innovative adaptation methodology for English LLMs, grounded in instruction-tuning data reflecting knowledge capabilities and safety values within a specific culture. Without the need for pre-training, this approach efficiently enables the short-term rapid adaptation of large models to the cultures of specific countries and regions. Evaluation results indicate that the English LLM, after being transformed with knowledge capabilities and safety values, knowledge and capabilities significantly enhance within specific domains, while fully aligning with the safety and values standards of the particular sociocultural context.




%
%
%
\bibliographystyle{splncs04}
\bibliography{main}
%




\end{document}